\documentclass[sigconf]{acmart}
\usepackage{booktabs} 
\usepackage{subfigure}
\usepackage{bm}

\usepackage{multirow}
\usepackage{algorithm}
\usepackage{algorithmic}
\usepackage{array}
\usepackage{graphicx}
\usepackage{listings}
\usepackage[np, autolanguage]{numprint}
\usepackage{enumitem}

\usepackage{soul}
\usepackage[skip=0pt]{caption}
\usepackage{subfigure}
\usepackage{amsmath}

\usepackage{xspace}
\usepackage{fancyhdr}

\newcommand{\paratitle}[1]{\vspace{1.5ex}\noindent\textbf{#1}}
\newcommand{\ie}{\emph{i.e.,}\xspace}

\newcommand{\eg}{\emph{e.g.,}\xspace}
\newcommand{\etal}{\emph{et al.}\xspace}
\newcommand{\ignore}[1]{}

\newcommand{\yscomment}[1]{\textcolor{blue}{[YS: #1]}}

\ignore{
  \AtBeginDocument{%
  \providecommand\BibTeX{{%
    \normalfont B\kern-0.5em{\scshape i\kern-0.25em b}\kern-0.8em\TeX}}}

\setcopyright{acmcopyright}
\copyrightyear{2018}
\acmYear{2018}
\acmDOI{10.1145/1122445.1122456}
}

\copyrightyear{2021} 
\acmYear{2021} 
\setcopyright{acmcopyright}\acmConference[WSDM '21]{Proceedings of the Fourteenth ACM International Conference on Web Search and Data Mining}{March 8--12, 2021}{Virtual Event, Israel}
\acmBooktitle{Proceedings of the Fourteenth ACM International Conference on Web Search and Data Mining (WSDM '21), March 8--12, 2021, Virtual Event, Israel}
\acmPrice{15.00}
\acmDOI{10.1145/3437963.3441753}
\acmISBN{978-1-4503-8297-7/21/03}

\begin{document}
\fancyhead{}
\fancyfoot{}

\title{Improving Multi-hop Knowledge Base Question Answering by Learning Intermediate Supervision Signals}

\author{Gaole He$^{1,4\dagger}$, Yunshi Lan$^{2}$, Jing Jiang$^{2}$, Wayne Xin Zhao$^{3,4*}$ and Ji-Rong Wen$^{1,3,4}$}\thanks{$^*$Corresponding author.\\ $\dagger$ This work is done when the first author visited SMU}
\affiliation{%
 \institution{$^1$School of Information, Renmin University of China}
 \institution{$^2$School of Information System, Singapore Management University}
  \institution{$^3$Gaoling School of Artificial Intelligence, Renmin University of China}
 \institution{$^4$Beijing Key Laboratory of Big Data Management and Analysis Methods}
}
\affiliation{%
  \institution{\{hegaole, jrwen\}@ruc.edu.cn, batmanfly@gmail.com, \{yslan, jingjiang\}@smu.edu.sg}
}	
\renewcommand{\shortauthors}{Gaole He, Yunshi Lan, Jing Jiang, Wayne Xin Zhao and Ji-Rong Wen}

\begin{abstract}
Multi-hop Knowledge Base Question Answering~(KBQA) aims to find the answer entities that are multiple hops away in the Knowledge Base (KB) from the entities in the question.
A major challenge is the lack of supervision signals at  intermediate steps.
Therefore, multi-hop  KBQA algorithms can only receive the feedback from the final answer, which makes the learning unstable or ineffective.

To address this challenge, we propose a novel teacher-student approach for the multi-hop KBQA task. In our approach, the student network aims to find the correct answer to the query, while the teacher network tries to learn intermediate supervision signals for improving the reasoning capacity of the student network. The major novelty lies in the design of the teacher network, where we utilize both forward and backward reasoning to enhance the learning of intermediate entity distributions.
By considering bidirectional reasoning, the teacher network can produce more reliable intermediate supervision signals, which can alleviate the issue of spurious reasoning.
Extensive experiments on three benchmark datasets have demonstrated the effectiveness of our approach on the KBQA task. The code to reproduce our analysis is available at \url{https://github.com/RichardHGL/WSDM2021_NSM}.
\end{abstract}

\ccsdesc[500]{Computing methodologies~Reasoning about belief and knowledge}
\ccsdesc[500]{Computing methodologies~Search with partial observations}
\ignore{
\ccsdesc[500]{Computing methodologies~Reasoning about belief and knowledge}

}

\keywords{Knowledge Base Question Answering; Teacher-student Network; Intermediate Supervision Signals}


\maketitle



\begin{figure}[htbp]
	\small
	\centering 
	\includegraphics[width=0.4\textwidth]{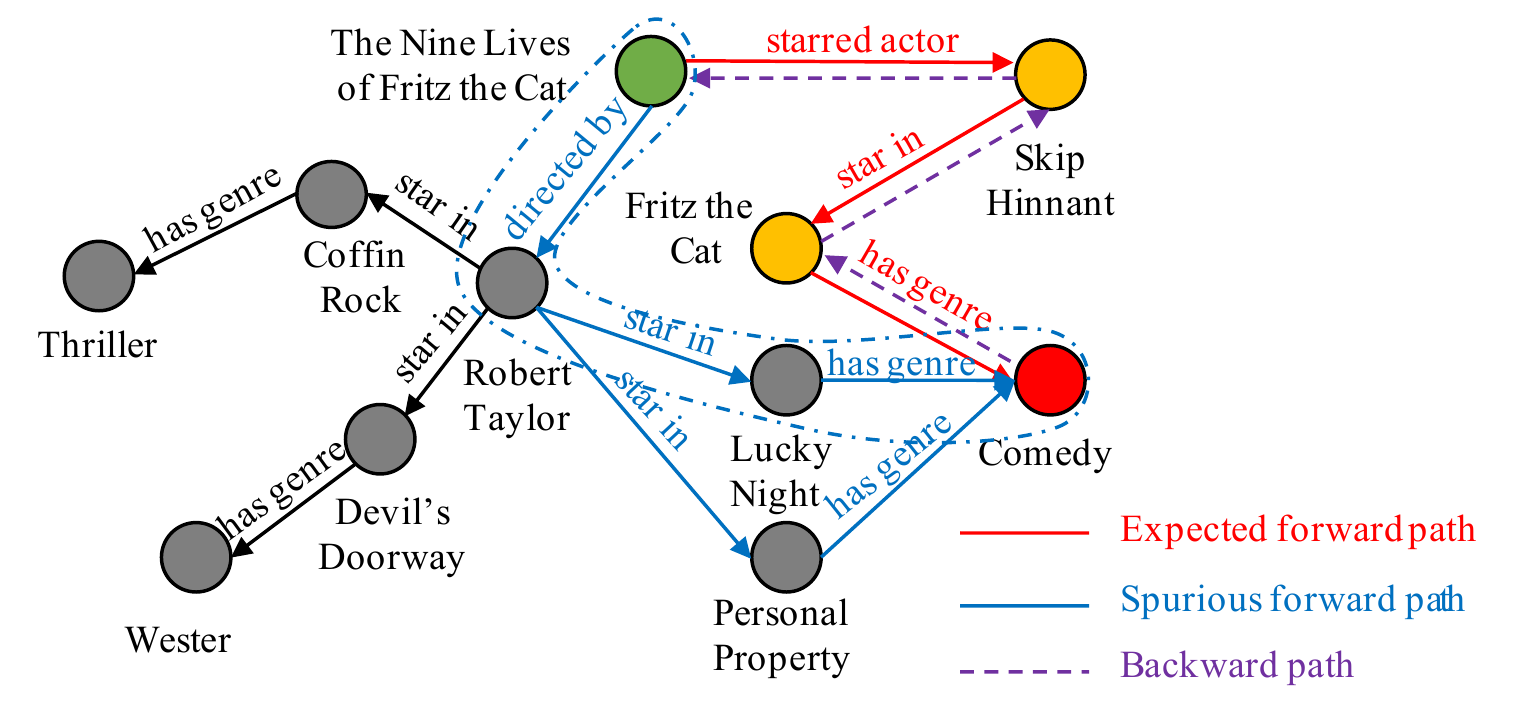}
	\caption{A spurious case from MetaQA-3hop dataset. We use green, red, yellow and grey circles to denote the topic entities, correct answer, intermediate entities and irrelevant entities respectively.} 
	\label{fig-spurious} 
\end{figure}	

\section{introduction}
Knowledge Base Question Answering~(KBQA) is a challenging task that aims at finding answers to questions expressed in natural language from a given knowledge base~(KB).
Traditional solutions~\cite{Ferrucci-AI-2010,Berant-EMNLP-2013,webqsp-ACL-2015,Dong-EMNLP-2015} usually develop a specialized pipeline consisting of multiple machine-learned or hand-crafted modules (\eg named entity recognition, entity linking). Recently, 
end-to-end deep neural networks~\cite{NSM-ACL-2017,GraftNet-EMNLP-2018} become the popular paradigm for this task by automatically learning data representations and network parameters.



For the KBQA task, there have been growing interests in solving complex questions that require a multi-hop reasoning procedure~\cite{Lan-ICDM-2019}, called \emph{multi-hop KBQA}. 
Besides the final answer, it is also important that a multi-hop KBQA algorithm can identify a reasonable relation path leading to the answer entities~\cite{SRN-WSDM-2020,Dua-ACL-2020}. 
In some cases,  even if the answer was correctly found, the relation path might be spurious. 
We present an example of spurious multi-hop reasoning in  Fig.~\ref{fig-spurious}. 
The question is ``\emph{what types are the films starred by actors in the nine lives of fritz the cat?}''. 
Besides the correct path (with red arrows),
two spurious paths  (with blue arrows) which include entities who are directors at the first step can also reach the correct answer.
It is mainly due to the lack of  supervision signals at the intermediate reasoning steps (which we call \emph{intermediate supervision signals}).
For the multi-hop KBQA task,  training data is typically in the form of $\langle question, answer \rangle$ instead of the ideal form of $\langle question, relation~path \rangle$. 
Therefore, multi-hop reasoning algorithms can only receive the feedback at the final answer using such datasets. 


\ignore{\textcolor{blue}{Challenges.}
When only weak supervision signal is provided, model is usually trained to conduct multi-hop reasoning without feedback along the intermediate steps. In other words, models get feedback for the whole reasoning process only at the end of reasoning through comprison between prediction and answers.
However, reasoning answers for a complex question typically require multi-hop reasoning on knowledge base.
Given feedback only at the end of reasoning, the correctness of intermediate steps is uncertain. A case of spurious reasoning in multi-hop KBQA is shown in Fig.~\ref{fig-spurious}.
This creates a main problem: spurious multi-hop reasoning in the training phase degrade the quality of training data. 
}

To address this issue, several studies formulate multi-hop KBQA as a reinforcement learning~(RL) task~\cite{MINERVA-ICLR-2018,Victoria-EMNLP-2018,SRN-WSDM-2020}. 
They set up a policy-based agent to sequentially extend its inference path until it reaches a target entity. 
Its states are usually defined as tuple of query and current entity, and action as traverse on KB through outgoing edges of current entity.
RL-based approaches heavily rely on the terminal reward to bias the search. 
To prevent spurious paths in the search, reward shaping~\cite{Victoria-EMNLP-2018,SRN-WSDM-2020} and action dropout~\cite{Victoria-EMNLP-2018} have been proposed to improve the model learning.
However, these solutions either require expert experience or still lack effective supervision signals at intermediate steps. 



\ignore{
\textcolor{blue}{main idea of previous methods.} Some previous work have tried to address it with heuristic techniques, such as reward shaping~\cite{Victoria-EMNLP-2018,SRN-WSDM-2020}, action dropout~\cite{Victoria-EMNLP-2018}. However, this practice may require expert knowledge about what requirement the intermediate steps or final answers should fullfill.
Some works (e.g., ~\cite{GraftNet-EMNLP-2018}) incorporate external knowledge (e.g. Wikipedia free text) to enrich the distributed representations of questions and KG constituents. However, this practice may not work when no external knowledge is available for a specific domain.
And most previous work try to provide feedback with respect to semantic matching between question in natural language and the reasoned relational path. But they don't estimate which entity may be qualified to be reasoned at intermediate steps.
}

Different from previous studies, our idea is to set up two models with different purposes for multi-hop KBQA.
The main model aims to find the correct answer to the query, while the auxiliary  model tries to learn intermediate 
supervision signals for improving the reasoning capacity of  the main model. Specifically, the auxiliary model infers which entities at the intermediate steps are more relevant to the question, and these entities are considered as intermediate supervision signals.
Although the idea is appealing, it is  challenging to learn an effective auxiliary model, since we do not have such labeled data for training.

\ignore{
Different from previous studies, our solution is to adopt the teacher-student learning paradigm for multi-hop KBQA. In our approach, the teacher and student  have different purposes.
The student aims to find the correct answer to the query, while the teacher tries to learn intermediate 
supervision signals for improving the reasoning capacity of  the student.  In specific, we  consider entity distributions  at each reasoning step inferred by the teacher  as intermediate 
supervision signals.
However, it is still challenging to learn such a teacher model that can accurately capture  intermediate entity distributions, since we do not have such labeled data for training.
}

Our solution is inspired by the bidirectional search algorithms (\eg bidirectional BFS~\cite{bbfs-IJCAI-1999}) on graphs, in which an ideal path connecting the source and the destination can be more effectively identified with bidirectional exploration. 
Indeed, for KBQA we also have two different views to consider the task setting: the forward reasoning that finds the path starting from the topic entities (\ie entities in the queries) to the answer entities and the backward reasoning that returns from answer entities to the topic entities.
Most existing methods only consider forward reasoning. 
However, it is possible to jointly model the two reasoning processes, since topic entities and answer entities are all known in the training data.
Such a bidirectional reasoning mechanism is able to incorporate additional self-supervision signals at intermediate steps. 
As shown in Fig.~\ref{fig-spurious}, the entity distribution obtained by  forward reasoning  at the second step should be similar to that from backward reasoning at the first step. 
Irrelevant entities ``\emph{Devil's Doorway}'' and ``\emph{Coffin Rock}'' are likely to be reached at the second reasoning step of forward reasoning but unreachable at the first step of backward reasoning. To maintain the correspondence between the two processes, we should avoid including the director ``\emph{Robert Taylor}'' at the first step of forward reasoning.
Such a potential correspondence is useful to improve the learning of each individual reasoning process at intermediate steps. That is the key point how we learn reliable  intermediate supervision signals.

To this end, in this paper, we propose a novel teacher-student approach for the multi-hop KBQA task. 
Specifically, the student network (\ie the main model), which aims to find the answer,  is implemented by adapting the Neural State Machine~(NSM)~\cite{NSM-NIPS-2019} from visual question answering. 
In our approach, the student network can improve itself according to intermediate entity distributions learned from the teacher network.  
The major novelty lies in the design of the teacher network (\ie the auxiliary model), which provides intermediate supervision signals. 
We utilize the correspondence between the state information from the forward and  backward reasoning processes to enhance the learning of intermediate entity distributions.  We further design two reasoning architectures that support the integration between  forward and backward reasoning.
By considering bidirectional reasoning, the teacher network can alleviate the issue of spurious reasoning, and produce more reliable intermediate supervision signals.  

To evaluate our approach, we conduct extensive experiments on three benchmark datasets. Extensive experiments have demonstrated the effectiveness of our approach on the multi-hop KBQA task, especially for cases lacking training data.
To the best of our knowledge, it is the first time that intermediate supervision signals have been explicitly learned with a teacher-student framework.

\section{Related Work}
Our work is closely related to the studies on KBQA, multi-hop reasoning and teacher-student framework.

\paratitle{Knowledge Base Question Answering.} For the KBQA task, various methods have been developed over the last decade. They can be categorized into two groups: semantic parsing based methods and retrieval based methods. Semantic parsing based methods~\cite{Berant-EMNLP-2013,webqsp-ACL-2015,Yih-ACL-2016,NSM-ACL-2017,Lan-ACL-2020} learn a semantic parser that converts natural language questions into intermediate logic forms, which can be executed against a KB. Retrieval-based methods~\cite{Dong-EMNLP-2015,KVMem-EMNLP-2016,Xu-ACL-2016,GraftNet-EMNLP-2018,PullNet-EMNLP-2019} directly retrieve answers from the KB in light of the information conveyed in the questions.

Recently, researchers pay more attention to multi-hop based KBQA. 
Some work~\cite{MetaQA-AAAI-2018, KVMem-EMNLP-2016, GraftNet-EMNLP-2018} employed classical methods (e.g., Variational Reasoning Network, Key-Value Memory Network and Graph Convolution Network) to conduct multi-hop reasoning within the KB.
Moreover, Sun \etal~\cite{PullNet-EMNLP-2019} and Saxena \etal~\cite{Saxena-ACL-2020} leveraged extra corpus and enriched knowledge graph embeddings to boost the performance of multi-hop KBQA.
However, these methods take the performance of final prediction as the only objective, which are vulnerable to the spurious examples.

\paratitle{Multi-hop Reasoning.} In recent years, multi-hop reasoning becomes a hot research topic for both computer vision and natural language processing domains. 
Min \etal ~\cite{DecompRC-ACL-2019} proposed to decompose complex queries into several 1-hop queries and solved them by turn. 
Hudson \etal~\cite{MAC-ICLR-2018} designed a novel recurrent Memory, Attention, and Composition (MAC) cell, which splits complex reasoning into a series of attention-based reasoning steps. 
Das \etal~\cite{MINERVA-ICLR-2018,Victoria-EMNLP-2018} conducted multi-hop reasoning on a graph under the reinforcement learning setting and treated every reasoning step as an edge transition on the graph.
Besides, there are quite a few studies that adopt Graph Neural Network (GNN)~\cite{GCN-ICLR-2017,GAT-ICLR-2018} to conduct explicit reasoning on graph structure~\cite{GraftNet-EMNLP-2018, Hu-ICCV-2019}. 

\paratitle{Teacher-student Framework.}
Knowledge distillation (KD) is introduced and generalized by early work~\cite{Hinton-Distill-2015}.
They proposed a teacher-student framework, where a complicated high-performance model and a light-weight model are treated as teacher and student respectively.
The predictions of the teacher model are treated as ``soft labels'' and the student model is trained to fit the soft labels. 
While knowledge distillation was initially proposed for model compression, recent work~\cite{Furlanello-ICML-2018,Zhang-CVPR-2018} found that applying the soft labels as the training target can help the student achieve better performance.

Several studies also apply the teacher-student framework in question answering task.
Yang \etal~\cite{Yang-WSDM-2020} designed a multi-teacher knowledge distillation paradigm in a Web Question Answering system.
Do \etal~\cite{Do-ICCV-2019} and Hu \etal~\cite{Hu-EMNLP-2018} applied the teacher-student framework to visual question answering task and reading comprehension task, respectively.
In this work, we try to address spurious reasoning caused by weak supervision in the multi-hop KBQA task with an elaborate teacher-student framework.

\section{Preliminary}
In this section, we introduce the background and define the task.

\paratitle{Knowledge Base~(KB)}. A knowledge base typically organizes factual information as a set of triples, denoted by $\mathcal{G} = \{\langle e,r,e' \rangle| e,e' \in \mathcal{E}, r \in \mathcal{R}\}$, where $\mathcal{E}$ and $\mathcal{R}$ denote the entity set and relation set, respectively. 
A triple $\langle e,r,e' \rangle$ denotes that relation $r$ exists between head entity $e$ and tail entity $e'$.
Furthermore, we introduce \emph{entity neighborhood} to denote the set of triples involving an entity $e$, denoted by $\mathcal{N}_e = \{\langle e, r, e' \rangle \in \mathcal{G} \} \cup \{\langle e', r, e \rangle \in \mathcal{G} \}$, 
containing both incoming and outgoing triples for $e$.
For simplicity, we replace a triple $\langle e, r, e' \rangle$ with its  reverse triple   $\langle e', r^{-1}, e \rangle$, so that we can have $\mathcal{N}_e=\{\langle e', r, e \rangle \in \mathcal{G} \}$.
For convenience, we further use italic bold fonts to denote the embeddings of entities or relations. 
Let $\bm{E} \in \mathbb{R}^{d \times |\mathcal{E}|}$ and $\bm{R} \in \mathbb{R}^{d \times |\mathcal{R}|}$ denote the embedding matrices  for entities and relations in the KB, respectively, and each column vector $\bm{e}\in \mathbb{R}^{d}$ or $\bm{r}\in \mathbb{R}^{d}$ is a $d$-dimensional embedding for entity $e$ or relation $r$.

\paratitle{Knowledge Base Question Answering~(KBQA)}. We focus on factoid question answering over a knowledge base. We assume that a KB $\mathcal{G}$ is given as the available resource and the answers will be the entities in $\mathcal{G}$. 
Formally, 
given a natural language question $q = \{w_1, w_2, ..., w_{l}\}$ and a KB $\mathcal{G}$, the task of KBQA is to figure out the answer entitie(s), denoted by the set  $\mathcal{A}_q$, to query $q$ from the candidate entity set $\mathcal{E}$. The entities mentioned in a question are  called \emph{topic entities}.
Specially, we consider solving complex questions where the answer entities are multiple hops away from the topic entities in the KB, called  \emph{multi-hop KBQA}. 
\section{The proposed approach}
In this section, we present the proposed approach for the multi-hop KBQA task under the teacher-student framework. 



\subsection{Overview}
A major difficulty for multi-hop KBQA is that it usually lacks supervision signals at intermediate reasoning steps, since only the answer entities are given as ground-truth information. To tackle this issue, we adopt the recently proposed teacher-student learning framework~\cite{Hinton-Distill-2015,FitNets-ICLR-2015}.
The main idea is to train a student network that focuses on the multi-hop KBQA task itself, while another teacher network is trained to provide (pseudo) supervision 
signals (\ie inferred entity distributions in our task) at  intermediate reasoning steps for improving the student network.

\ignore{We design an elaborative approach under teacher-student framework to labeling and utilizing intermediate entities for the multi-hop KBQA task. The teacher model try to combine forward reasoning with backward reasoning and label intermediate steps through providing intermediate entity distributions. Considering the difference between forward reasoning and backward reasoning, teacher model may introduce noise to prevent it from obtaining optimal performance for multi-hop KBQA task.
For that sake, we utilize intermediate entity distributions labeled by teacher model with a separate student model.
}

In our approach, the student network is implemented based on Neural State Machine~(NSM)~\cite{NSM-NIPS-2019}, which was originally proposed for visual question answering on scene graph extracted from image data.
We adapt it to the multi-hop KBQA task by considering KB as a graph, and maintain a gradually learned entity distribution over entities during the multi-hop reasoning process. 
To develop the teacher network, we modify the architecture of  
NSM by incorporating a novel bidirectional reasoning mechanism, so that it can learn more reliable  entity distributions at intermediate reasoning steps, which will be subsequently used by the student network as the supervision signals.

\ignore{
To conduct multi-hop reasoning on KB, we set up two components with different purposes, namely instruction component and reasoning component. 
At every reasoning step, the instruction component generate instruction vector to guide the reasoning component, while the reasoning component seek for intermediate entities under the guidance of instruction vector. Through multi-step cooperation of both components, we decompose the complex reasoning process into simple reasoning steps, and every step try to find intermediate answers for part of original complex question. For a reasoning process of $N$ steps, we obtain answers for the whole question from the final predicted distribution.
}


In what follows, we first describe the adapted architecture of NSM for multi-hop KBQA, and then present the teacher network and model learning. 



\ignore{
\begin{figure*}[ht]
 \centering
  \subfigure[Overview of student model (neural state machine).]{\label{fig-student}
  \centering
  \includegraphics[width=0.48\textwidth]{pic/student.pdf}
 }
  \subfigure[A example of teacher with concatenated structure for 3-hop reasoning.]{\label{fig-teacher-2}
  \centering
  \includegraphics[width=0.48\textwidth]{pic/teacher-2-new.pdf}
 }
 \centering
 \caption{Models}
 \label{fig-model}
\end{figure*}	
}

\subsection{Neural State Machine for Multi-hop KBQA}
\label{sec-nsm}
We present an overall sketch of NSM in Fig.~\ref{fig-student}. It mainly consists  of an instruction component and a reasoning component. The instruction component sends instruction vectors to the reasoning component, while the reasoning component infers the entity distribution and learns the entity representations.  
\ignore{
\begin{figure*}[tbp]
 \centering
  \subfigure[Overview of teacher-student framework.]{\label{fig-overview}
  \centering
  \includegraphics[width=0.48\textwidth]{pic/overview.pdf}
 }
  \subfigure[Illustration of the two reasoning steps for neural state machine on question ``\emph{which person directed the movies starred by john krasinski}?''. In different reasoning steps, the instruction vector focuses on different parts of the question.]{\label{fig-student}
  \centering
  \includegraphics[width=0.48\textwidth]{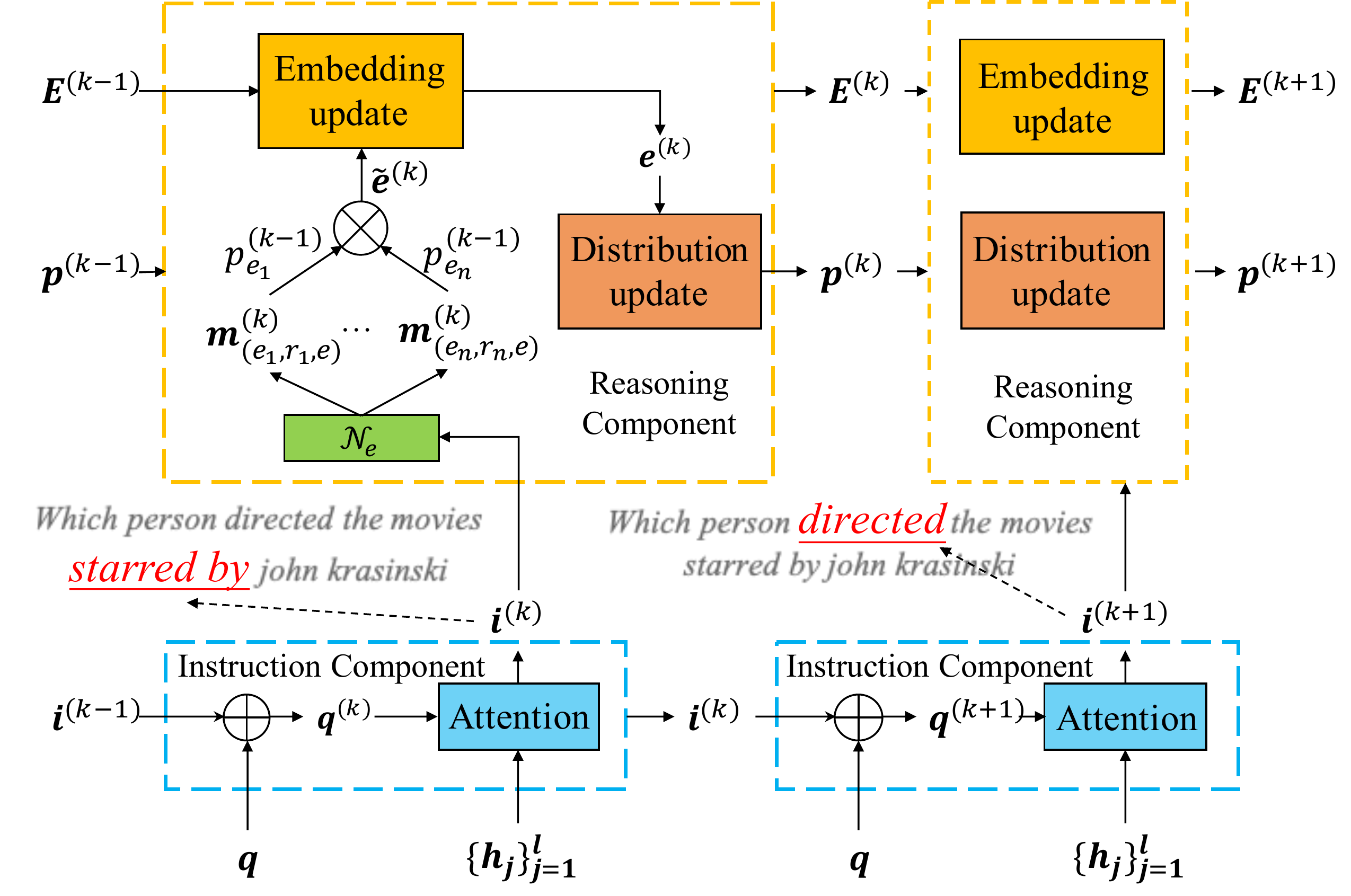}
 }
 \centering
 \caption{Models}
 \label{fig-model}
\end{figure*}		
}

\begin{figure}[htbp]
	\small
	\centering 
	\includegraphics[width=0.48\textwidth]{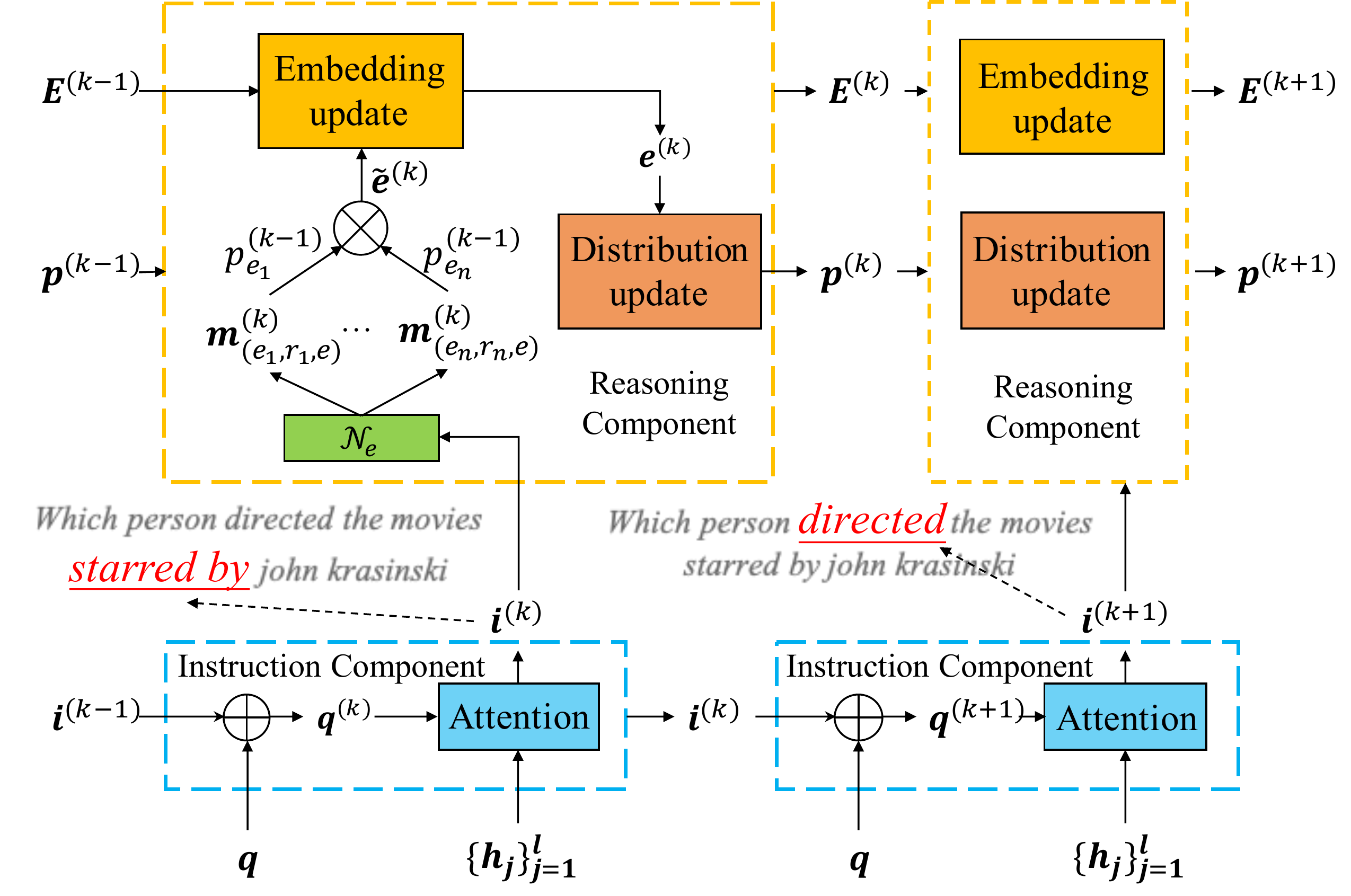}
	\caption{Illustration of the two reasoning steps for neural state machine on question ``\emph{which person directed the movies starred by john krasinski?}''. In different reasoning steps, the instruction vector focuses on different parts of the question.} 
	\label{fig-student} 
\end{figure}

\subsubsection{Instruction Component}
\label{sec-message}
We first describe how to transform a given natural language question into a series of instruction vectors that control the reasoning process.
The input of the instruction component consists of a query embedding and an instruction vector from the previous reasoning step. 
The initial instruction vector is set as zero vector.
We utilize GloVe~\cite{glove-emnlp-2014} to obtain the embeddings of the query words. Then we adopt a standard LSTM encoder to obtain a set of hidden states $\{\bm{h}_j \}_{j=1}^{l}$, where $\bm{h}_j \in \mathbb{R}^d$ and $l$ is the length of the query. 
After that, the last hidden state is considered to be the question representation, \ie $\bm{q}=\bm{h}_l$. 
Let $\bm{i}^{(k)} \in \mathbb{R}^d$ denote the instruction vector at the $k$-th reasoning step.
We adopt the following method to learn the instruction vector $\bm{i}^{(k)}$:
\begin{equation}
\begin{split}
\bm{i}^{(k)}	&= \sum_{j=1}^l\alpha^{(k)}_{j}\bm{h}_j,\\
\alpha^{(k)}_{j}	&= \text{softmax}_j \big(\bm{W}_{\alpha}(\bm{q}^{(k)}\odot\bm{h}_j) + \bm{b}_{\alpha}\big),\\
\bm{q}^{(k)}	&=	\bm{W}^{(k)}[\bm{i}^{(k-1)}; \bm{q}] + \bm{b}^{(k)},
\end{split}
\end{equation}	
where $\bm{W}^{(k)} \in \mathbb{R}^{d \times 2d}, \bm{W}_{\alpha}  \in \mathbb{R}^{d \times d}$ and $\bm{b}^{(k)}, \bm{b}_{\alpha} \in \mathbb{R}^{d}$ are parameters to learn.
The core idea is to attend to specific parts of a query when learning the instruction vectors at different time steps. In such a process, we also dynamically update the query representation, so that it can incorporate the information of previous instruction vectors. By repeating the process above, we can obtain a list of instruction vectors $\{ \bm{i}^{(k)} \}_{k=1}^{n}$ after $n$ reasoning steps. 

\subsubsection{Reasoning Component}
\label{sec-graph-transition}

\ignore{
\textcolor{blue}{
This section is inspired mainly from NSM~\citep{NSM-NIPS-2019,Pathcon-arxiv-2020} model.
}	
}

Once we obtain the instruction vector $\bm{i}^{(k)}$, we can use it as a guiding signal for the reasoning component. 
The input of the reasoning component consists of the instruction vector of the current step, and the entity distribution and entity embeddings obtained from the previous reasoning step.
The  output of the reasoning component includes the entity distribution $\bm{p}^{(k)}$ and the entity embeddings $\{\bm{e}^{(k)}\}$.
First, we set the initial entity embeddings by considering 
the  relations involving $e$: 
\begin{equation}
\label{eq-init-gnn}
\bm{e}^{(0)} = \sigma\bigg(\sum_{\langle e', r, e \rangle \in \mathcal{N}_e }\bm{r} \cdot \bm{W}_T\bigg),
\end{equation}
where $\bm{W}_T \in \mathbb{R}^{d \times d}$ are the parameters to learn. Unlike previous studies~\cite{KVMem-EMNLP-2016, GraftNet-EMNLP-2018}, we explicitly utilize the information of related relation types for encoding entities. 
In the multi-hop KBQA task, a reasoning path consisting of multiple relation types can reflect important semantics that lead to the answer entities. Besides, such a method is also useful to reduce the influence of noisy entities, and easy to apply to unseen entities of known context relations.
Note that we do not use the original embedding of $e$ when initializing $\bm{e}^{(0)}$ because for intermediate entities along the reasoning path the identifiers of these entities are not important; it is the relations that these intermediate entities are involved in that matter the most.

Given a triple $\langle e', r, e \rangle$, a match vector $\bm{m}^{(k)}_{\langle e', r, e \rangle}$ is learned by 
 matching  the current instruction $\bm{i}^{(k)}$ with relation vector $r$:
\begin{equation}
\bm{m}^{(k)}_{\langle e', r, e \rangle}  =  \sigma\big(\bm{i}^{(k)}\odot\bm{W}_{R}\bm{r}\big),
\end{equation}
where $\bm{W}_R \in \mathbb{R}^{d \times d}$ are the parameters to learn.
Furthermore, we aggregate the matching messages from neighboring triples and assign weights to them according to how much attention they receive at the last reasoning step:
\ignore{
To obtain reasoning in graph context, we consider both message carried by neighborhood edges and how much attention neighborhood entities obtain at last reasoning steps.
}
\begin{equation}
\tilde{\bm{e}}^{(k)}  =  \sum_{\langle e', r, e \rangle \in \mathcal{N}_e}\bm{p}^{(k-1)}_{e'}\cdot \bm{m}^{(k)}_{\langle e', r, e \rangle},
\end{equation}
where $\bm{p}^{(k-1)}_{e'}$ is the assigned probability of entity $e'$ at the last reasoning step, which we will explain below.
Such a representation is able to capture the relation semantics associated with an entity in the KB. 
Then, we update entity embeddings as follows:
\begin{equation}
\begin{split}
\label{eq-st-update}
\bm{e}^{(k)} &= \text{FFN}([\bm{e}^{(k-1)}; \tilde{\bm{e}}^{(k)}]),
\end{split}
\end{equation}
where $\text{FFN}(\cdot)$ is a feed-forward layer taking as input of both previous embedding $\bm{e}^{(k-1)}$ and relation-aggregated embedding $\tilde{\bm{e}}^{(k)}$.


Through such a process,  both the relation path (from topic entities to answer entities) and its matching degree with the question can be encoded into node embeddings.
The probability distribution over intermediate entities derived at step $k$ can be calculated as:
\begin{equation}
\label{eq-prob-st}
\bm{p}^{(k)} = \text{softmax} \big( \bm{E}^{(k)^T} \bm{w} \big),
\end{equation}
where $\bm{E}^{(k)}$ is a matrix where each column vector is the embedding of an entity at the $k$-th step, and $\bm{w} \in \mathbb{R}^{d}$ are the parameters that derive the entity distribution $\bm{p}^{(k)}$, and $\bm{E}^{(k)}$ is the updated entity embedding matrix by Eq.~\ref{eq-st-update}.

\subsubsection{Discussion}
For our task,  
the reason that we adopt the NSM model as the student network are twofold. 
First, our core idea is to utilize intermediate entity distributions derived from the teacher network as the supervision signals for the student network.
In contrast, most previous multi-hop KBQA methods do not explicitly maintain and learn such an entity distribution at intermediate steps. Second, NSM can be considered as a special graph neural network, which has excellent reasoning capacity over the given knowledge graph.  As shown in Section 4.2.2,  the learning of entity distributions and entity embeddings can indeed correspond to the general ``\emph{propagate-then-aggregate}'' update mechanism of graph neural networks. We would like to utilize such a powerful neural architecture to solve the current task.  

\ignore{
To provide labeling of intermediate entities, the basic model is supposed to provide both intermediate entity distribution and answer distribution. By unifing them in NSM framework, the intermediate entity distribution can be viewed as answer distribution for part of the original complex question. 
Previous work GraftNet~\cite{GraftNet-EMNLP-2018} adopt personalized pagerank (PPR) to control the GNN update and conduct multi-hop reasoning on question-specific graph. However, GraftNet don't unify such PPR with prediction of answers, and the PPR of GraftNet don't obtain any supervision signal through reasoning process. So we don't adopt GraftNet as our basic model.
}

The NSM~\cite{NSM-NIPS-2019} was proposed to conduct visual reasoning in an abstract latent space. We make two major adaptations for  multi-hop KBQA. First, in Eq.~\ref{eq-init-gnn}, we initialize the node embeddings by aggregating the embeddings of those relations involving the entity.  In our task, the given KB is usually very large. An entity is likely to be linked to a large number of other entities.
Our initialization method is able to reduce the influence of noisy entities, focusing on the important relational semantics. Besides, it is also easy to generalize to new or unseen entities with known relations, which is especially important to incremental training. 
Second, in Eq.~\ref{eq-st-update}, we update entity embeddings by integrating previous embedding $\bm{e}^{(k-1)}$ and relation-aggregated embedding $\tilde{\bm{e}}^{(k)}$. 
For comparison, original NSM~\cite{NSM-NIPS-2019} separately modeled the two parts, whereas we combine the two factors in a unified update procedure, which is useful to derive more effective node embeddings.

\begin{figure*}[htbp]
 \centering

\subfigure[Illustration of 3-hop parallel reasoning.]{\label{fig-teacher-1}
  \centering
  \includegraphics[width=0.48\textwidth]{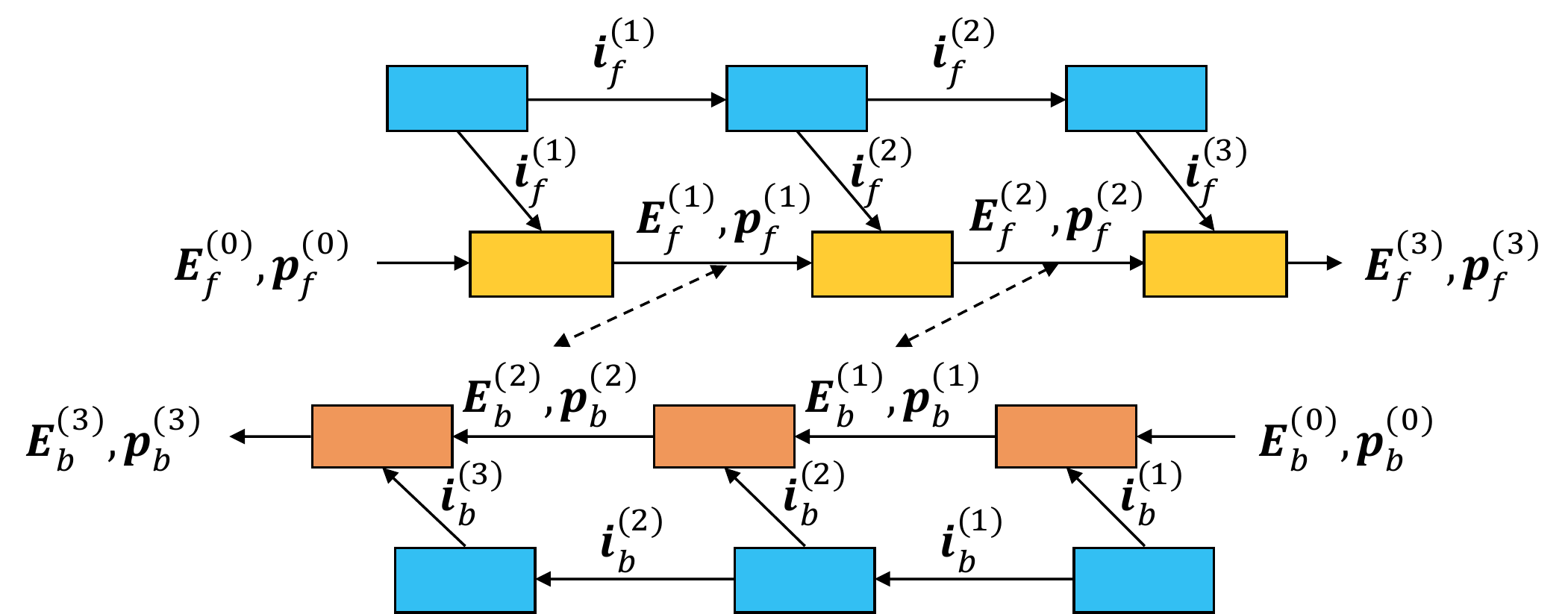}
 }
 \subfigure[Illustration of 3-hop hybrid reasoning.]{\label{fig-teacher-2}
  \centering
  \includegraphics[width=0.48\textwidth]{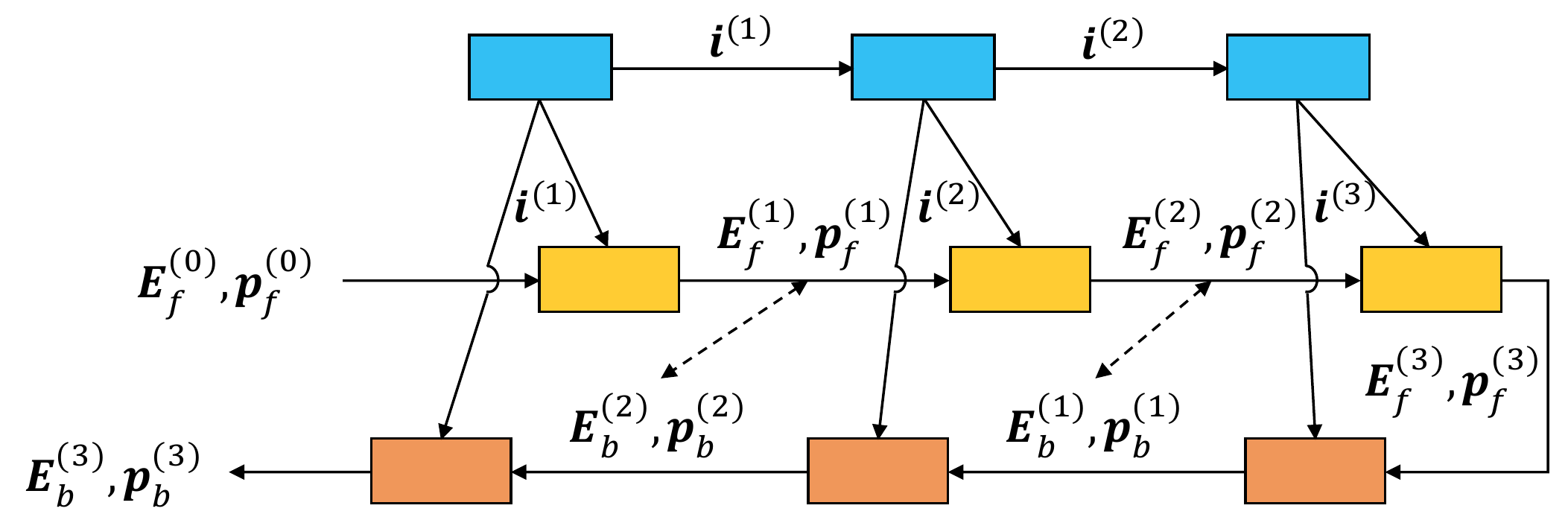}
 }
 \centering
 \caption{Illustration of the designed teacher architectures. We use blue, yellow and orange squares to denote the instruction component, forward reasoning component and backward reasoning component, respectively. The dotted arrows link the corresponding intermediate entity distributions of the two reasoning processes. 
 We use $f$ and $b$ as subscripts to distinguish forward reasoning and backward reasoning, respectively.}
 \label{fig-teacher}
\end{figure*} 

\subsection{The Teacher Network}
Different from the student network, the teacher network aims to learn or infer reliable   entity distributions at intermediate reasoning steps. 
Note that there are no such labeled entity distributions for training the teacher network. 
Instead, inspired by the bidirectional search algorithm (\eg bidirectional BFS~\cite{bbfs-IJCAI-1999}), we incorporate the bidirectional reasoning mechanism for enhancing the learning of intermediate entity distributions in the teacher network.

\subsubsection{Bidirectional Reasoning for Multi-hop KBQA}
Given a knowledge base, the reasoning process for multi-hop KBQA can be considered to be an exploration and search problem on the graph. 
Most existing multi-hop KBQA methods start from the topic entities and then look for the possible answer entities, called \emph{forward reasoning}. 
On the other hand, the opposite search from answer entities to topic entities (which we refer to as \emph{backward reasoning}) has been neglected by previous studies.
Our core idea is to consider the exploration in both directions and let the two reasoning processes synchronize with each other at intermediate steps. 
In this way, the derived intermediate entity distributions can be more reliable than those learned from a single direction. 
More specifically, given a $n$-hop reasoning path, let $\bm{p}_f^{(k)}$ and $\bm{p}_b^{(n-k)}$ denote the entity distributions from the forward reasoning at the $k$-th step and from the backward reasoning at the $(n-k)$-th step, respectively.
The key point is that the two distributions should be similar or consistent if the two reasoning processes have been stable and accurate, \ie $\bm{p}_f^{(k)} \approx \bm{p}_b^{(n-k)}$. We will utilize such a correspondence as constraints in the following models. 

\ignore{
For multi-hop KBQA, there should be only one correct reasoning, if we ask human to label intermediate answers given both seed entities and final answers. If we try to reason a question within $N$ steps, the intermediate distribution predicted by forward reasoning at step $k$ is supposed to be consistent with intermediate distribution predicted by backward reasoning at step $N-k$.
\begin{equation}
\bm{p}_f^{(k)} \approx \bm{p}_b^{(N-k)}, k=1, ..., N - 1
\end{equation}
}	

\ignore{
To encourage such consistent prediction, we add regularization term about entity distributions of corresponding steps to loss.
\begin{equation}
\mathcal{L}_c = \sum_{i=1}^{N-1} D_{JS}(p_f^{(i)}, p_b^{(N-i)})
\end{equation}	
}

\subsubsection{Reasoning Architectures}
\label{sec-teacher}
Based on the idea above, we design two kinds of neural architectures for the teacher network, namely parallel reasoning and hybrid  reasoning. 

\paratitle{Parallel Reasoning}. The first way is to set up two separate NSMs for both forward and backward reasoning, respectively. 
These two NSM networks are relatively isolated, and do not share any parameters.
We only consider incorporating correspondence constraints on the intermediate entity distributions between them. 

\paratitle{Hybrid Reasoning}. In the second way, we share the same instruction component and arrange the  two reasoning processes in a cycled pipeline. 
Besides the correspondence constraints, the two processes receive the same instruction vectors. 
Furthermore, the derived information at the final step of the forward reasoning is fed into the backward reasoning as initial values. 
Formally,  the following equations hold in this case: 
\begin{equation}
\begin{split}
\bm{p}^{(0)}_b &= \bm{p}^{(n)}_f,~\bm{E}^{(0)}_b = \bm{E}^{(n)}_f,\\
\bm{i}^{(k)}_b &= \bm{i}^{(n+1-k)}_f,  k=1,...,n.\\
\end{split}
\end{equation}

We present the illustrative examples of the parallel reasoning and hybrid reasoning in Fig.~\ref{fig-teacher-1} and Fig.~\ref{fig-teacher-2}.
Comparing the two reasoning architectures, it can be seen that parallel reasoning has a more loose integration, while hybrid reasoning requires a deeper 
fusion between the information from both reasoning processes.  
Unlike bidirectional  BFS, in our task, backward reasoning might not be able to  exactly mimic the inverse process of forward reasoning, since the two processes correspond to different semantics in multi-hop KBQA. Considering this issue, we share the instruction vectors and recycle the final state of the forward reasoning for initializing backward reasoning. In this way, backward reasoning receives more information about forward reasoning, so that it can better trace back the 
reasoning path of forward reasoning. 

\subsection{Learning with the Teacher-Student Framework}
In this part, we present the details of model learning with our teacher-student framework.

\subsubsection{Optimizing the Teacher Network}
The two reasoning architectures of the teacher network can be optimized in the same way. 
We mainly consider 
two parts of loss, namely reasoning loss and correspondence loss.  

The reasoning loss reflects the capacity of predicting the accurate entities, which can be decomposed into two directions: 
\begin{equation}
\begin{split}
\mathcal{L}_f &= D_{KL}\big(\bm{p}_f^{(n)}, \bm{p}_f^{*}\big),~\mathcal{L}_b = D_{KL}\big(\bm{p}_b^{(n)}, \bm{p}_b^{*}\big),\\
\end{split}
\end{equation}
where $\bm{p}_f^{(n)}$ ($\bm{p}_b^{(n)}$) denotes the final entity distribution for forward (backward) reasoning process, $\bm{p}_f^{*}$ ($\bm{p}_b^{*}$) denotes the ground-truth entity distribution, and $D_{KL}(\cdot, \cdot)$ is the Kullback-Leibler divergence~\cite{kldiv-1951}, which measures the difference between the two distributions in an asymmetric way. To obtain $\bm{p}_f^{*}$ and $\bm{p}_b^{*}$, we transform the occurrences of ground-truth entities into a frequency-normalized distribution. 
Specifically, if $k$ entities in the graph are ground-truth entities, they are assigned a  probability of $\frac{1}{k}$  in the final distribution.

The correspondence loss reflects the consistency degree between intermediate  entity distributions from the two reasoning processes. 
It can be computed by summing the loss at each intermediate step:
\begin{equation}
\label{eq-loss-corr}
\mathcal{L}_c = \sum_{k=1}^{n-1} D_{JS}\big(\bm{p}_f^{(k)}, \bm{p}_b^{(n-k)}\big),
\end{equation}
where  $D_{JS}(\cdot, \cdot)$ is the Jensen-Shannon divergence~\cite{jsdiv}, which measures the difference between two distributions in a symmetric way.

To combine the above loss terms, we define the entire loss function of the teacher network $\mathcal{L}_t$ as:
\begin{equation}
\label{eq-loss-teahcer}
\mathcal{L}_t = \mathcal{L}_f + \lambda_b\mathcal{L}_b + \lambda_c \mathcal{L}_c,
\end{equation}
where $\lambda_b \in (0, 1)$ and $\lambda_c \in (0, 1)$ are hyper-parameters to control the weights of the factors.

\subsubsection{Optimizing the Student Network}
After the teacher model is trained to convergence, we can obtain intermediate entity distributions in the two 
reasoning processes of the teacher network. 
We take the average of the two distributions as the supervision signal:
\begin{equation}
\begin{split}
\bm{p}_t^{(k)} &= \frac{1}{2}\big(\bm{p}_f^{(k)}+\bm{p}_b^{(n-k)}\big), ~~~~~k=1, ... , n-1\\
\end{split}
\end{equation}

As described before, we adopt the NSM model as the student network to conduct forward reasoning.
Besides the reasoning loss, we also incorporate the loss  between the predictions of the student network and the supervision signal of the teacher network: \begin{equation}
\label{eq-loss-student}
\begin{split}
\mathcal{L}_1  &= D_{KL}(\bm{p}_s^{(n)}, \bm{p}_f^{*}),\\
\mathcal{L}_2  &= \sum_{k=1}^{n-1}D_{KL}(\bm{p}_s^{(k)}, \bm{p}_t^{(k)}),\\
\mathcal{L}_s &= \mathcal{L}_1 + \lambda  \mathcal{L}_2.\\
\end{split} 
\end{equation}
where $\bm{p}_t^{(k)}$ and  $\bm{p}_s^{(k)}$ denote  the intermediate entity distributions at the $k$-th step from the teacher network and student network, respectively, and $\lambda$ is a hyperparameter to tune.

In practice, labeled data for intermediate reasoning steps is seldom available. 
Most existing methods only rely on the final answer to learn the entire model, which may not be well trained or form spurious reasoning paths. 
Our approach adopts the teacher network  for improving the student network. 
The main novelty is to utilize both forward and backward reasoning in producing more reliable  intermediate entity distributions.
Note that we do not incorporate any additional labeled data for training intermediate reasoning steps in the teacher network. 
Instead, we try to learn such intermediate entity distributions by enforcing the correspondence in the bidirectional reasoning process. 
To our knowledge, backward reasoning has been seldom considered in multi-hop KBQA task, especially its correspondence with forward reasoning. Such an idea is indeed related to recent progress in self-supervised learning~\cite{ssl-tpami-2020}, in which we leverage internal supervision signal to learn the model.

\ignore{
Although there have been a few studies which conduct multi-hop reasoning for KBQA task, our approach has two major differences. 
First, our approach conduct multi-hop reasoning and pay attention to both the correctness of final prediction and entities involved in intermediate steps. As a comparison, most of previous methods only focus on the correctness of final prediction.
Second, our approach take both forward reasoning and backward reasoning into consideration, and provide intermediate labeling for entities based on combined view. As a comparison, backward reasoning has been seldom considered in previous methods for KBQA task.
To our knowledge, it is the first time that backward reasoning and labeling of intermediate reasoning steps have been utilized for multi-hop KBQA task.
}
\section{Experiment}
In this section, we perform the evaluation experiments for our approach on the KBQA task. 

\subsection{Datasets}

We adopt three benchmark datasets for the multi-hop KBQA task:

\textbf{MetaQA}~\citep{MetaQA-AAAI-2018} contains more than 400k single and multi-hop (up to 3-hop) questions in the domain of movie, containing three 
datasets, namely MetaQA-1hop, MetaQA-2hop and MetaQA-3hop.



\ignore{
\paratitle{MetaQA}~\citep{MetaQA-AAAI-2018} contains more than 400k single and multi-hop (up to 3-hop) questions in the domain of movies. The questions were constructed using the knowledge base provided with the WikiMovies~\citep{KVMem-EMNLP-2016} dataset. We use the “vanilla” version of the queries\footnote{For this version, the 1-hop questions in MetaQA are exactly the same as WikiMovies.}. We use the KB supplied with the WikiMovies dataset, and use exact match on surface forms to perform entity linking. The KB used here is relatively small, with about 43k entities and 135k triples. MetaQA datasets contain 3 datasets, namely MetaQA-1hop, MetaQA-2hop and MetaQA-3hop.
}

\textbf{WebQuestionsSP (webqsp)}~\citep{webqsp-ACL-2015} contains 4737 natural language questions that are answerable using Freebase as the knowledge base. The questions require up to 2-hop reasoning from knowledge base. We use the same train/dev/test splits as GraftNet~\citep{GraftNet-EMNLP-2018}.

\textbf{Complex WebQuestions 1.1 (CWQ)}~\citep{CWQ-NAACL-2018} is generated from WebQuestionsSP by extending the question entities or adding constraints to answers. There are four types of question: composition (45$\%$), conjunction (45$\%$), comparative (5$\%$), and superlative (5$\%$). The questions require up to 4-hops of reasoning on the KB. 
\ignore{
\begin{table}[!h]
	\centering
	\caption{Statistics of all datasets. $\#$entity denotes average number of entities in subgraph, and coverage denotes the ratio of at least one answer in subgraph.}
	\label{tab:datasets}%
	\begin{footnotesize}
	\begin{tabular}{l | r r r | r r r}
		\hline
			Datasets		&	Train&	Dev&	Test&	$\#$entity&	recall& coverage\\
		\hline
		MetaQA-1hop	&	96,106&	9,992&	 9,947&	487.6&	99.9$\%$&	100$\%$\\
		MetaQA-2hop	&	118,980&	14,872&	 14,872&	469.8&	99.9$\%$&	100$\%$\\
		MetaQA-3hop	&	114,196&	14,274&	 14,274&	497.9&	83.1$\%$&	99.0$\%$\\
		webqsp	&	 2,848&	 250&	  1,639&	1429.8&	93.5$\%$&	94.9$\%$\\
		CWQ	&	27,639&	3,519&	  3,531&	1305.8&	77.5$\%$&	79.3$\%$\\
		\hline
	\end{tabular}%
	\end{footnotesize}
\end{table}%
}
\begin{table}[!h]
	\centering
	\caption{Statistics of all datasets. ``$\#$entity'' denotes average number of entities in subgraph, and ``coverage'' denotes the ratio of at least one answer in subgraph.}
	\label{tab:datasets}%
	\begin{tabular}{l | r r r | r r}
		\hline
			Datasets		&	Train&	Dev&	Test&	$\#$entity& coverage\\
		\hline
		MetaQA-1hop	&	96,106&	9,992&	 9,947&	487.6&	100$\%$\\
		MetaQA-2hop	&	118,980&	14,872&	 14,872&	469.8&	100$\%$\\
		MetaQA-3hop	&	114,196&	14,274&	 14,274&	497.9&	99.0$\%$\\
		webqsp	&	 2,848&	 250&	  1,639&	1,429.8&	94.9$\%$\\
		CWQ	&	27,639&	3,519&	  3,531&	1,305.8&	79.3$\%$\\
		\hline
	\end{tabular}%
\end{table}%
Following~\citep{GraftNet-EMNLP-2018,PullNet-EMNLP-2019}, we use the topic entities labeled in original datasets and adopt PageRank-Nibble algorithm~(PRN) \citep{PPR-Andersen-2006} to find KB entities close to them.
With these entities, we can obtain a relatively small subgraph that is likely to contain the answer entity.
For CWQ and webqsp datasets, we first obtain the neighborhood graph within two hops of topic entities and then run PRN algorithm on it.  We further expand one hop for CVT entities in Freebase to obtain the neighborhood subgraph. As shown in  Table~\ref{tab:datasets},  2-hop graphs are sufficient to cover most of the answer entities.  While on MetaQA datasets, we run PRN algorithm on the entire KB. 
Specifically, we use the PRN algorithm~\citep{PPR-Andersen-2006} with $\epsilon=1e^{-6}$ and then select the $m$ top-scoring entities. We set $m = 500$ for the smaller MetaQA KB and $m = 2000$ for larger Freebase. 
For the reserved triples, both their head and tail entities are obtained from the top $m$ entities identified by PRN algorithm.
We summarize the statistics of the three datasets in Table~\ref{tab:datasets}.

\subsection{Experimental Setting}

\subsubsection{Evaluation Protocol.} 
We follow~\cite{GraftNet-EMNLP-2018,PullNet-EMNLP-2019} to cast the multi-hop KBQA task as a ranking task for evaluation.
For each test question in a dataset, a list of answers are returned by a model according to their predictive probabilities. We adopt two evaluation metrics widely used in previous works, namely \emph{Hits@1} and \emph{F1}.  Specifically, Hits@1 refers to whether the top answer is correct.
For all the methods, we learn them using the training set, and optimize the parameters using the validation set and compare their performance on the test set.

\subsubsection{Methods to Compare.} We consider the following methods for performance comparison:

    $\bullet$ \textbf{KV-Mem}~\cite{KVMem-EMNLP-2016} maintains a memory table for retrieval, which stores KB facts encoded into key-value pairs.
	

	$\bullet$ \textbf{GraftNet}~\cite{GraftNet-EMNLP-2018} adopts a variant of graph convolution network to perform multi-hop reasoning on heterogeneous graph.
	
	$\bullet$ \textbf{PullNet}~\cite{PullNet-EMNLP-2019} utilizes the shortest path as supervision to train graph retrieval module and conduct multi-hop reasoning with GraftNet on the retrieved sub-graph.
	
	$\bullet$ \textbf{SRN}~\cite{SRN-WSDM-2020} is a multi-hop reasoning model under the RL setting, which solves multi-hop question answering through extending inference paths on knowledge base.
	
	$\bullet$ \textbf{EmbedKGQA}~\cite{Saxena-ACL-2020} conducts multi-hop reasoning through matching pretrained entity embedings with question embedding obtained from RoBERTa~\cite{RoBERTa-Liu-2019}.
	
	$\bullet$ \textbf{NSM}, \textbf{NSM}$_{+p}$ and \textbf{NSM}$_{+h}$ are three variants of our model, which (1) do not use the teacher network, (2) use the teacher network with parallel reasoning, and (3) use the teacher network with hybrid reasoning, respectively.

\subsubsection{Implementation Details}
Before training the student network, we pre-train the teacher network on multi-hop KBQA task. To avoid overfitting, we adopt early-stopping by evaluating Hits@1 on the validation set every 5 epochs. We optimize all models with Adam optimizer, where the batch size is set to 40. The learning rate is tuned amongst \{0.01, 0.005, 0.001, 0.0005, 0.0001\}. The reasoning steps is set to $4$ for CWQ dataset, while $3$ for other datasets. 
The coefficient $\lambda$ (in Eq.~\ref{eq-loss-student}) and $\lambda_b, \lambda_c$(in Eq.~\ref{eq-loss-teahcer}) are tuned amongst \{0.01, 0.05, 0.1, 0.5, 1.0\}.

\subsection{Results}

The results of different methods for KBQA are presented in Table~\ref{tab:res}. 
It can be observed that:

(1) Among the baselines, KV-Mem performs the worst.
This is probably because it does not explicitly consider the complex 
reasoning steps.
Most methods perform very well on the MetaQA-1hop and MetaQA-2hop datasets, which require only up to 2 hops of reasoning.
On the other hand, the other datasets seem to be more difficult, especially the webqsp and CWQ datasets. 
Overall, EmbedKGQA and PullNet are better than the other baselines. 
PullNet trains an effective subgraph retrieval module based
on the shortest path between topic entities and answer entities. 
Such a module is specially useful to reduce the subgraph size and produce high-quality candidate entities.

(2) Our base model (\ie the single student network) NSM performs better than the competitive baselines in most cases. It is developed based on a graph neural network with two novel extensions for this task (Sec.~\ref{sec-nsm}).  
The gains of teacher-student framework show variance on different datasets. 
Specifically, on the two most difficult datasets, namely Webqsp and CWQ, the variants of NSM$_{+p}$ and NSM$_{+h}$ are substantially better than NSM and other baselines. 
These results have shown the effectiveness of the teacher network in our approach, which largely improves the student network. 
Different from SRN and PullNet, our approach designs a novel bidirectional reasoning mechanism to learn more reliable intermediate supervision signals.
Comparing NSM$_{+p}$ and NSM$_{+h}$, we find that their results are similar. On Webqsp and CWQ datasets, the hybrid reasoning is slightly better to improve the student network than parallel reasoning.

\begin{table}[htbp]
	\centering
	\caption{Performance comparison of different methods for KBQA (Hits@1 in percent). We copy the results for KV-Mem, GraftNet and PullNet from~\cite{PullNet-EMNLP-2019}, and copy the results for SRN and EmbedKGQA from \cite{SRN-WSDM-2020,Saxena-ACL-2020}. Bold and underline fonts denote the best and the second best methods.}
	\label{tab:res}%
	\begin{small}
		\begin{tabular}{m{0.07\textwidth} c c c c c c}
			\hline
			Models& Webqsp& MetaQA-1& MetaQA-2 & MetaQA-3 & CWQ\\
			\cline{2-6}
			\hline
			KV-Mem&	46.7& 96.2& 82.7& 48.9& 21.1\\
			GraftNet &	66.4& 97.0& 94.8& 77.7& 32.8\\
			PullNet &	68.1& 97.0& \textbf{99.9}& 91.4& 45.9\\
			SRN &	-& 97.0& 95.1& 75.2& -\\
			EmbedKGQA&	66.6& \textbf{97.5}& \underline{98.8}& \underline{94.8}& -\\
			\hline
			NSM	&	68.7& 97.1& \textbf{99.9}& \textbf{98.9}&	47.6\\
			\hline
			NSM$_{+p}$&	\underline{73.9}& \underline{97.3}& \textbf{99.9}& \textbf{98.9}& \underline{48.3}	\\
			NSM$_{+h}$&	\textbf{74.3}& 97.2& \textbf{99.9}& \textbf{98.9}& \textbf{48.8}\\
			\hline
		\end{tabular}%
	\end{small}
\end{table}%

\subsection{Detailed Performance Analysis}
Table~\ref{tab:res} has shown that our approach overall has a better performance.
Next, we perform a series of detailed analysis experiments. For clarity, we only incorporate the results of NSM as the reference, since it performs generally well among all the baselines.

\ignore{
, our basic model NSM and our proposed approach shows a better overall performance than the baselines. Here, we zoom into the results and check whether our approach is indeed better than baselines in specific cases. For ease of comprison, we only incorporate the results of NSM as the reference, since it perform generally well among all the baselines.
}

\subsubsection{Ablation Study}
Previous experiments have indicated that the major improvement is from the contribution of the teacher network. Here, we compare the effect of different implementations of the teacher network. 
The compared variants include: (1) \underline{$\emph{NSM}_{+f}$} using only the forward reasoning (unidirectional); (2)  \underline{$\emph{NSM}_{+b}$} using only the backward reasoning (unidirectional); (3) \underline{$\emph{NSM}_{+p}$} using the parallel reasoning (bidirectional); (4) \underline{$\emph{NSM}_{+h}$} using the hybrid reasoning (bidirectional); (5) \underline{$\emph{NSM}_{+p,-c}$} removing the correspondence loss (Eq.~\ref{eq-loss-corr}) from $\emph{NSM}_{+p}$; and (6) \underline{$\emph{NSM}_{+h,-c}$} removing the correspondence loss (Eq.~\ref{eq-loss-corr}) from $\emph{NSM}_{+h}$.  
In Table~\ref{tab:ablation}, we can see that  unidirectional reasoning is consistently worse than bidirectional reasoning: the variants of $\text{NSM}_{+f}$ and $\text{NSM}_{+b}$ have a lower performance than the other variants. 
Such an observation verifies our assumption that bidirectional reasoning can improve the learning of intermediate supervision signals. 
Besides, by removing the correspondence loss from the teacher network, the performance substantially drops, which indicates that forward and backward reasoning can mutually  enhance each other. 

\begin{table}[htbp]
	\centering
	\caption{Ablation study of the teacher network (in percent).}
	\label{tab:ablation}%
		\begin{tabular}{c c c | c c}
			\hline
			\multirow{2}{*}{Models}&\multicolumn{2}{c}{Webqsp}&\multicolumn{2}{c}{CWQ}\\
			\cline{2-5}
			&Hits&F1&Hits&F1\\
			\hline
			NSM	&	68.7& 62.8&47.6&	42.4\\
			\hline
			$\text{NSM}_{+f}$&	70.7& 64.7& 47.2& 41.5\\
			$\text{NSM}_{+b}$&	71.1& 65.4& 47.1& 42.7\\
			$\text{NSM}_{+p,-c}$&	72.5&	66.5& 47.7& 42.7\\
			$\text{NSM}_{+h,-c}$&	73.0& 	66.9& 47.5& 42.1\\
			$\text{NSM}_{+p}$&	73.9& 66.2	&48.3&\textbf{44.0}\\
			$\text{NSM}_{+h}$&	\textbf{74.3}& \textbf{67.4}& \textbf{48.8} & \textbf{44.0}\\
			\hline
		\end{tabular}%
\end{table}%

\begin{figure}[htbp]
 \centering
 \subfigure[Varying $\lambda$ on webqsp dataset.]{\label{fig-varing-webqsp}
  \centering
  \includegraphics[width=0.225\textwidth, height=0.2\textwidth]{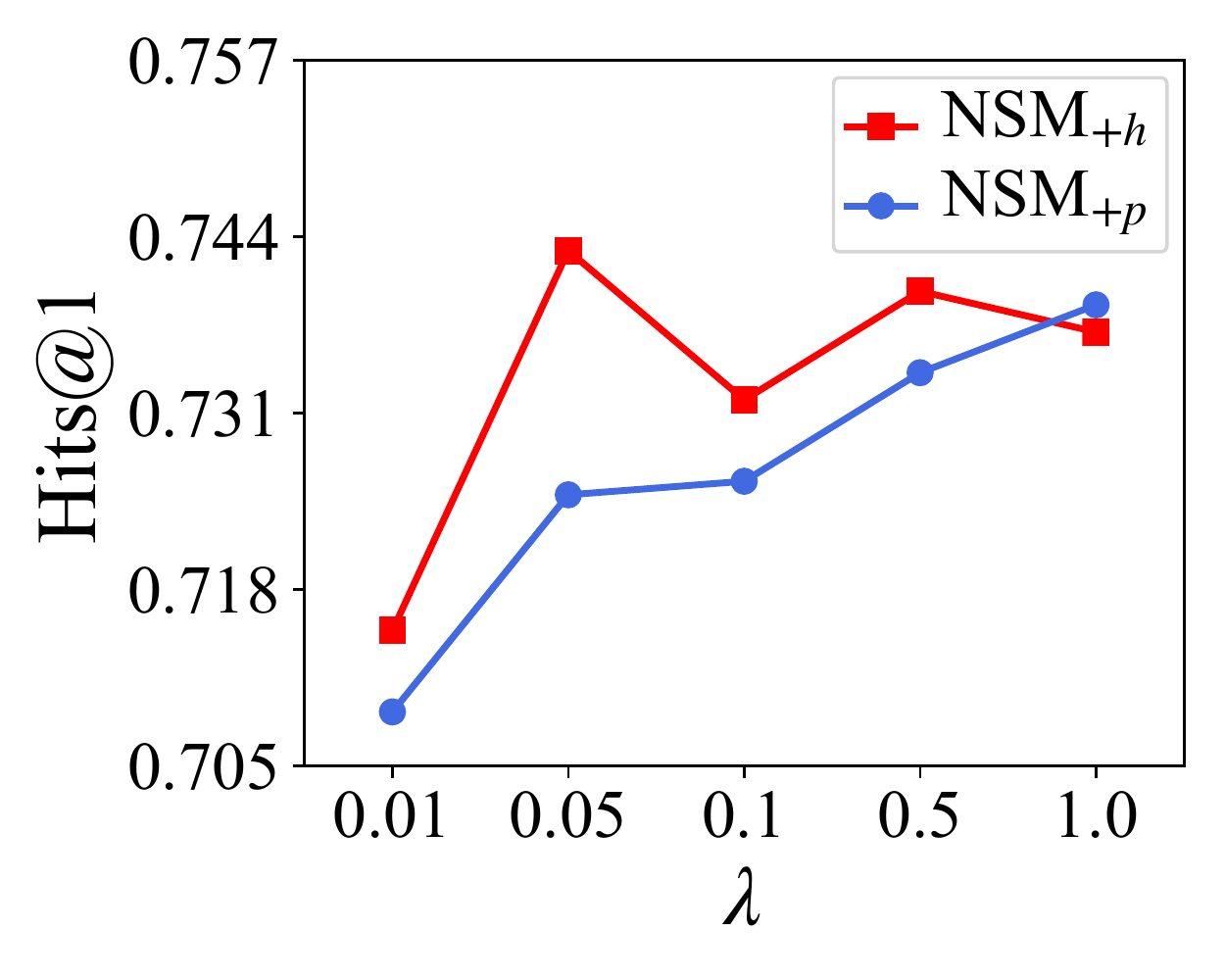}
 }
 \subfigure[Varying $\lambda$ on CWQ dataset.]{\label{fig-varing-CWQ}
  \centering
  \includegraphics[width=0.225\textwidth, height=0.2\textwidth]{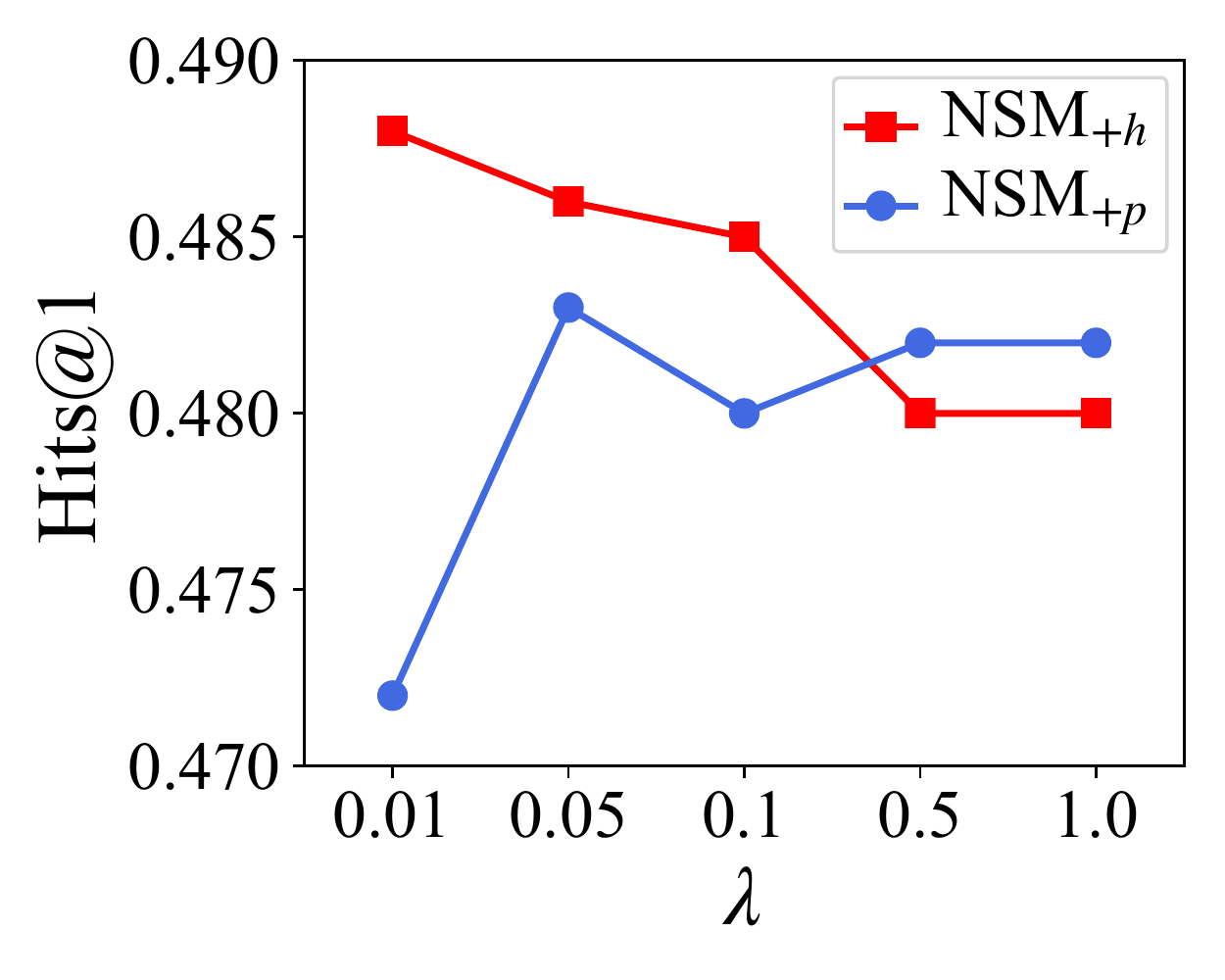}
 }
 \centering
 \caption{Performance tuning of our approach.}
 \label{fig-parameter-tuning}
\end{figure}

\begin{figure*}[htbp]
 \centering
  \subfigure[The student network before improvement.]{\label{fig-case-nsm}
  \centering
  \includegraphics[width=0.32\textwidth]{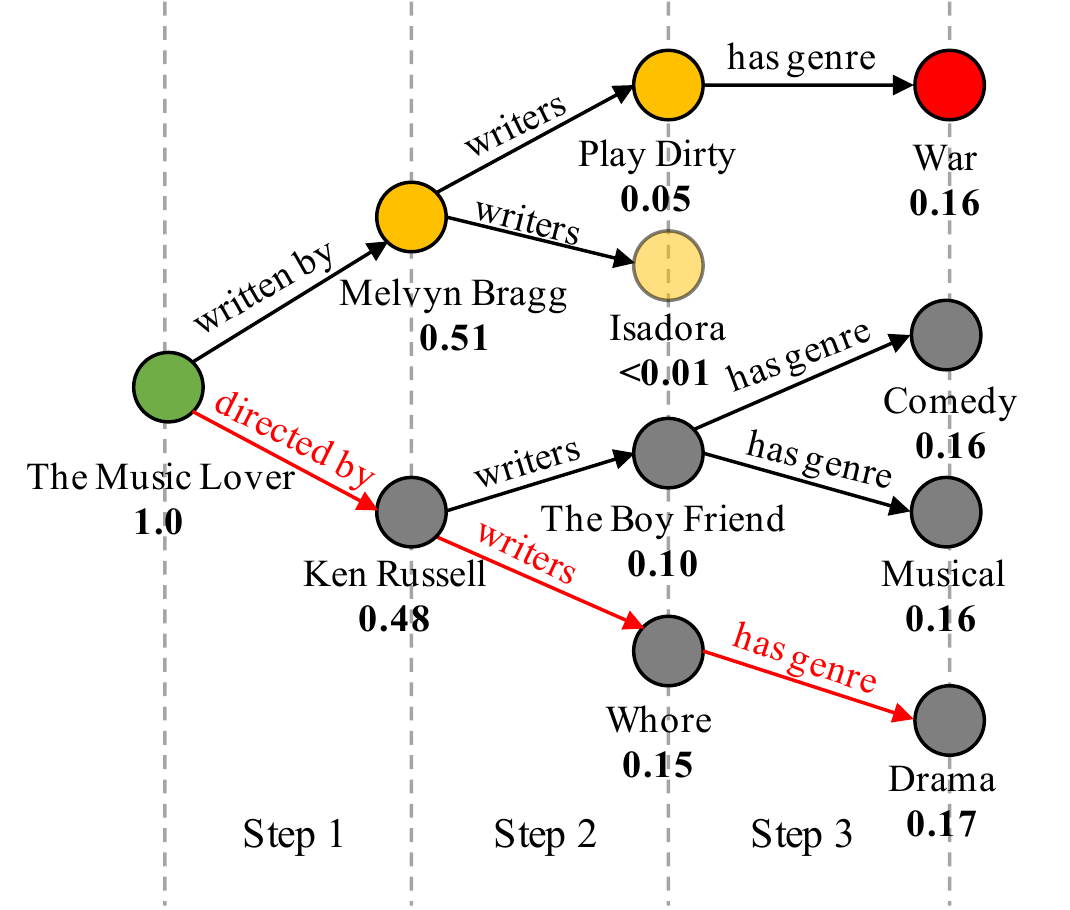}
 }
  \subfigure[The teacher network with hybrid reasoning.]{\label{fig-case-teacher}
  \centering
  \includegraphics[width=0.32\textwidth]{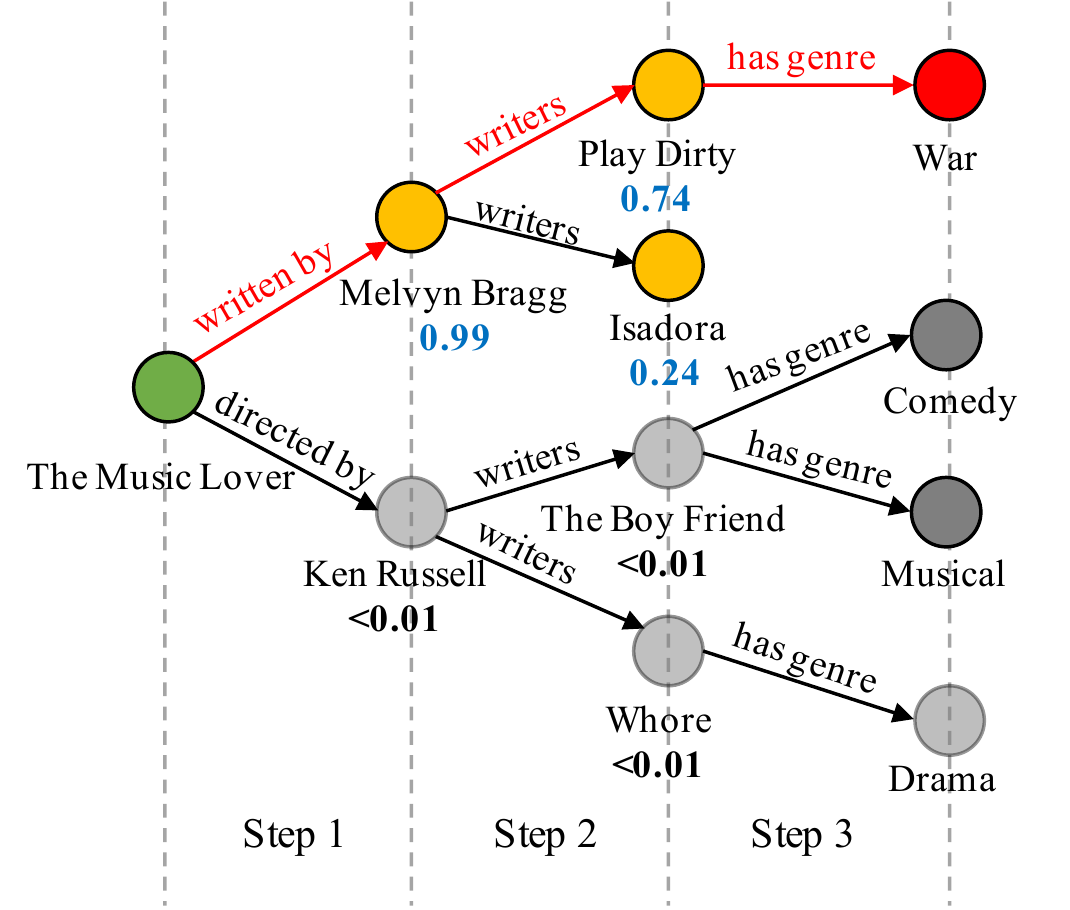}
 }
 \subfigure[The student network after improvement.]{\label{fig-case-student}
  \centering
  \includegraphics[width=0.32\textwidth]{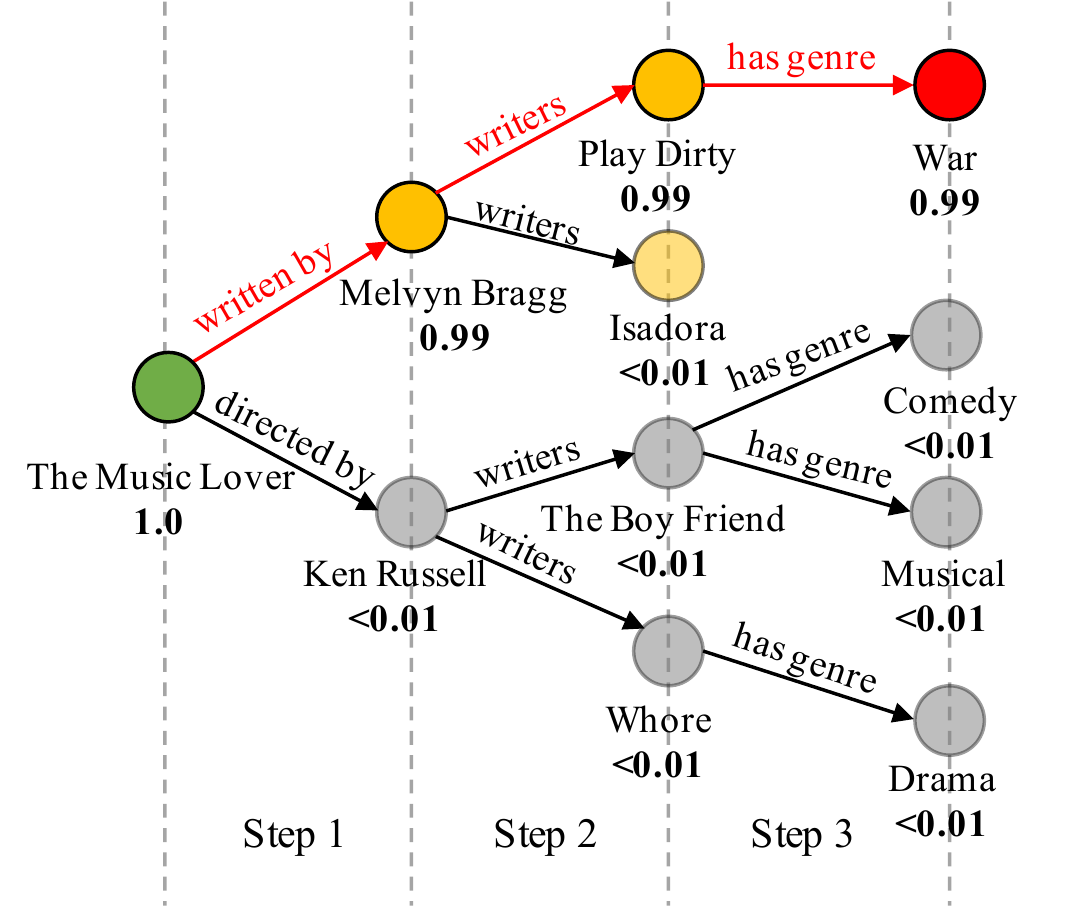}
 }
 \centering
 \caption{A case from the MetaQA-3hop dataset. We use green, red, yellow and grey circles to denote the topic entity, correct answer, intermediate entities and irrelevant entities respectively. The red colored edges denote the actual  reasoning paths for different methods. The color darkness indicates the relevance degree of an entity by a method. For simplicity, we only visualize the entities with a probability equal to or above $0.01$.
 }
 \label{fig-case-study}
\end{figure*}

\subsubsection{Parameter Tuning} 
In our approach, we have several combination coefficients to tune, including $\lambda$ in Eq.~\ref{eq-loss-student}, and $\lambda_b$ and $\lambda_c$ in Eq.~\ref{eq-loss-teahcer}. We first tune $\lambda$ amongst \{0.01, 0.05, 0.1, 0.5, 1.0\}, which controls the influence of the teacher network on the student network. 
As shown in Fig.~\ref{fig-parameter-tuning},  hybrid reasoning seems to work
well with small $\lambda$ (\eg 0.05), while parallel reasoning works better with relatively large $\lambda$ (\eg 1.0). 
Similarly, we can tune the parameters of $\lambda_b$ and $\lambda_c$.
Overall, we find that  $\lambda_c=0.01$  and $\lambda_b=0.1$ are good choices for our approach. Another parameter to tune is the embedding dimension $d$ (which is set to 100), and we do not observe significant improvement when $d>100$. 
The reasoning steps $n$ should be adjusted for different datasets. We observe that our approach achieves the best performance on CWQ dataset with $n=4$, while $n=3$ for the other datasets with exhaustive search. Due to space limit, we omit these tuning results. 

\subsubsection{Evaluating Intermediate Entities}
A major assumption we made is that our teacher network can obtain more reliable intermediate entities than the student network. 
Here, we compare the performance of the two networks in finding intermediate entities. Since the MetaQA-3hop dataset is created using pre-defined templates, we can recover the ground-truth entities at intermediate hops. We consider it a retrieval task and adopt the standard \emph{Precision}, \emph{Recall} and \emph{F1} as evaluation metrics.  
From Table~\ref{tab:acc-int}, we can see that the teacher network is much better than the student network in finding intermediate entities, but has slightly worse performance at the second hop. 
Note that the results of the third hop have been omitted, since it is the last hop. 
Since the student network only utilizes forward reasoning, the results of the first hop are more important than those of subsequent hops. These results also explain why our teacher-student approach is better than the single student model. 

\begin{table}[htbp]
	\centering
	\caption{Performance comparison \emph{w.r.t.} different hops on MetaQA-3hop dataset (in percent).}
	\label{tab:acc-int}%
		\begin{tabular}{c c c c| c c c}
			\hline
			\multirow{2}{*}{Models}&\multicolumn{3}{c}{Hop 1}&\multicolumn{3}{c}{Hop 2}\\
			\cline{2-7}
			& Pre& Rec& F1& Pre& Rec& F1\\
			\hline
			Student&	61.0& 60.6& 60.4& \textbf{99.9}& 70.2& \textbf{80.8}\\
			Teacher$_{+p}$&	80.0& 59.0& 66.3& 95.0& 68.9& 78.8\\
			Teacher$_{+h}$&	\textbf{99.9}& 56.0& \textbf{70.9}& 99.7& 63.0& 75.4\\
			\hline
		\end{tabular}%
\end{table}%

\subsubsection{One-Shot Evaluation}
In Table~\ref{tab:res}, we have found that the improvement of our approach over the basic NSM model is very small on the MetaQA datasets. We suspect that this is because the amount of training data for MetaQA is more than sufficient:  $100K$ training cases for 
no more than 300 templates in each dataset. 
To examine this, we randomly sample a single training case for every question template from the original training set, which forms a  one-shot training dataset. 
We evaluate the performance of our approach trained with this new training dataset. The results are shown in Table~\ref{tab:res-1shot}. As we can see, our approach still works very well, and the improvement over the basic NSM becomes more substantial.

\ignore{
As MetaQA datasets only contain very limited question template: no more than 300 templates for every dataset, but around 10k training cases. 
Our basic model may be good enough to conduct accurate multi-hop reasoning under so many training cases.
To verify the effectiveness of our approach on MetaQA dataset, we randomly sample only 1 training case for every question template in training set to replace original training set, and keep the validation and testing set same as original. The results are shown in Table~\ref{tab:res-1shot}. These results indicate that our approach are useful to improve the performance under extremely sparse setting.
}

\begin{table}[htbp]
	\centering
	\caption{Results under one-shot setting (in percent).}
	\label{tab:res-1shot}%
		\begin{tabular}{c c c |c c |c c }
			\hline
			\multirow{2}{*}{Models}&\multicolumn{2}{c}{MetaQA-1}&\multicolumn{2}{c}{MetaQA-2}&\multicolumn{2}{c}{MetaQA-3}\\
			\cline{2-7}
			&Hits&F1&Hits&F1&Hits&F1\\
			\hline
			NSM	&	 93.3& 92.6 &  97.7& 96.0& 90.6& 74.5\\
			\hline
			NSM$_{+p}$&	 \textbf{94.3}&  \textbf{93.9}&  \textbf{98.7}& \textbf{96.4}& \textbf{97.0}& 79.8\\
			NSM$_{+h}$&	 93.9&  93.7&  98.4& 95.8& 95.6& \textbf{81.6}\\
			\hline
		\end{tabular}%
\end{table}%

\subsection{Case Study}


The major novelty of our approach lies in the teacher network.   Next, we present a case study for illustrating how it helps the student network.

Given the question ``\emph{what types are the movies written by the screenwriter of the music lovers}'', the correct reasoning path is 
 ``\emph{The Music Lovers}''~(movie) $\rightarrow_{\textit{written by}}$  ``\emph{Melvyn Bragg}''~(screenwriter) $\rightarrow_{\textit{write}}$ ``\emph{Play Dirty}''~(movie) $\rightarrow_{\textit{has genre}}$ ``\emph{War}''~(genre). Note that ``\emph{Isadora}'' is also qualified at the second step.
However, its genre is missing in the KB.
Fig.~\ref{fig-case-study} presents a comparison between the learned results of the student before improvement (\ie without the teacher network), the teacher network and the student network after improvement. 


As shown in Fig.~\ref{fig-case-nsm}, the original student network has selected a wrong path leading to an irrelevant entity. At the first hop, NSM mainly focuses on the two  entities ``\emph{Ken Russell}'' and ``\emph{Melvyn Bragg}''  with probabilities of $0.48$ and $0.51$ respectively. 
Since it mistakenly includes ``\emph{Ken Russell}'' (director of ``\emph{The Music Lovers}'') at the first reasoning step, it finally ranks ``\emph{Drama}'' as the top entity and chooses an irrelevant entity as the answer.
In comparison, the teacher network (Fig.~\ref{fig-case-teacher}) is able to combine forward and backward reasoning to enhance the intermediate entity distributions. As we can see, our teacher assigns a very high  probability of $0.99$ to the entity ``\emph{Melvyn Bragg}'' at the first step.
When the supervision signals of the teacher are incorporated into the student, it correctly finds the answer entity ``\emph{War}'' with a high probability of 0.99 (Fig.~\ref{fig-case-student}).

This example has shown that our teacher network indeed provides very useful supervision signals at intermediate steps to improve the student network. 

\section{Conclusion}
\ignore{
In this paper, we developed an elaborate approach based on teacher-student framework for the  multi-hop KBQA task. In our approach, the role of  the teacher network is to learn intermediate supervision signals to improve the student network. For this purpose,  
we utilized the correspondence between state information from a forward and a backward reasoning process to enhance the learning of intermediate entity distributions. 
We further designed two reasoning architectures that support the integration between forward and backward reasoning. We designed a generic neural state machine as student model for multi-hop KBQA task.
We conducted evaluation experiments with three benchmark datasets. The results show that our proposed model is superior to previous methods in terms of effectiveness for the multi-hop KBQA task. 

Currently, we adopt the NSM model as the student network. 
It is flexible to extend our approach to other  neural architectures  or learning strategies on graphs. In the future, we will also consider enhancing the entity embeddings  using KB embedding methods, and obtain better intermediate supervision signals.
}

In this paper, we developed an elaborate approach based on teacher-student framework for the multi-hop KBQA task. In our approach, the student network implemented by a generic neural state machine focuses on the task itself, while the teacher network aims to learn intermediate supervision signals to improve the student network. 
For the teacher network, we utilized the correspondence between state information from a forward and a backward reasoning process to enhance the learning of intermediate entity distributions. 
We further designed two reasoning architectures that support the integration between forward and backward reasoning. 
We conducted evaluation experiments with three benchmark datasets. The results show that our proposed model is superior to previous methods in terms of effectiveness for the multi-hop KBQA task.

Currently, we adopt the NSM model as the student network. 
It is flexible to extend our approach to other  neural architectures  or learning strategies on graphs. In the future, we will also consider enhancing the entity embeddings  using KB embedding methods, and obtain better intermediate supervision signals.



\section*{Acknowledgement}
We thank Kun Zhou and Junyi Li for the helpful  discussions.
This work is partially supported by the National Research Foundation, Singapore under its International Research Centres in Singapore Funding Initiative, the National Natural Science Foundation of China under Grant No. 61872369 and 61832017,   Beijing Academy of Artificial Intelligence~(BAAI), and Beijing Outstanding Young Scientist Program under Grant No. BJJWZYJH012019100020098. Any opinions, findings and conclusions or recommendations expressed in this material are those of the author(s) and do not reflect the views of National Research Foundation, Singapore. 
\bibliographystyle{ACM-Reference-Format}%
\bibliography{nsm}

\end{document}